\documentclass[conference]{IEEEtran}
\IEEEoverridecommandlockouts
\usepackage[utf8]{inputenc}
\usepackage{newunicodechar}
\usepackage[english]{babel}
\usepackage{amsmath,amssymb,amsfonts}
\usepackage{algorithmic}
\usepackage{graphicx}
\usepackage{textcomp}
\usepackage{xcolor}
\usepackage{tikz}
\usepackage{pgfplots}
\usepackage{diagbox}
\usepackage{colortbl}
\usepackage{multirow}
\usepackage{hhline}
\usepackage{array}
\usepackage{cite}
\usepackage{hyperref}
\def\BibTeX{{\rm B\kern-.05em{\sc i\kern-.025em b}\kern-.08em
    T\kern-.1667em\lower.7ex\hbox{E}\kern-.125emX}}
\graphicspath{ {./images/} }

\pgfplotsset{compat=1.17}
\IEEEoverridecommandlockouts
\IEEEpubid{\makebox[\columnwidth]{978-1-6654-4067-7/21/\$31.00~\copyright2021 IEEE\hfill} \hspace{\columnsep}\makebox[\columnwidth]{ }}

\newcommand\copyrighttext{%
  \footnotesize \textcopyright 2021 IEEE. Personal use of this material is permitted. Permission from IEEE must be obtained for all other uses, in any current or future media, including reprinting/republishing this material for advertising or promotional purposes, creating new collective works, for resale or redistribution to servers or lists, or reuse of any copyrighted component of this work in other works.
  DOI: \href{https://doi.org/10.1109/IEMTRONICS52119.2021.9422541}{10.1109/IEMTRONICS52119.2021.9422541}}
\newcommand\copyrightnotice{%
\begin{tikzpicture}[remember picture,overlay]
\node[anchor=south,yshift=10pt] at (current page.south) {\fbox{\parbox{\dimexpr\textwidth-\fboxsep-\fboxrule\relax}{\copyrighttext}}};
\end{tikzpicture}%
}

\title{Image Classification with CondenseNeXt for ARM-Based Computing Platforms\\}

\begin{document}
\author{\IEEEauthorblockN{Priyank Kalgaonkar}
\IEEEauthorblockA{\textit{Department of Electrical and Computer Engineering} \\
\textit{Purdue School of Engineering and Technology}\\
Indianapolis, Indiana 46202, USA. \\
pkalgaon@purdue.edu}
\and
\IEEEauthorblockN{Mohamed El-Sharkawy}
\IEEEauthorblockA{\textit{Department of Electrical and Computer Engineering} \\
\textit{Purdue School of Engineering and Technology}\\
Indianapolis, Indiana 46202, USA. \\
melshark@purdue.edu}
}

\maketitle
\copyrightnotice
\IEEEpubidadjcol
\begin{abstract}
In this paper, we demonstrate the implementation of our ultra-efficient deep convolutional neural network architecture: CondenseNeXt on NXP BlueBox, an autonomous driving development platform developed for self-driving vehicles. We show that CondenseNeXt is remarkably efficient in terms of FLOPs, designed for ARM-based embedded computing platforms with limited computational resources and can perform image classification without the need of a CUDA enabled GPU. CondenseNeXt utilizes the state-of-the-art depthwise separable convolution and model compression techniques to achieve a remarkable computational efficiency.

Extensive analyses are conducted on CIFAR-10, CIFAR-100 and ImageNet datasets to verify the performance of CondenseNeXt Convolutional Neural Network (CNN) architecture. It achieves state-of-the-art image classification performance on three benchmark datasets including CIFAR-10 (4.79\% top-1 error), CIFAR-100 (21.98\% top-1 error) and ImageNet (7.91\% single model, single crop top-5 error). CondenseNeXt achieves final trained model size improvement of 2.9+ MB and up to 59.98\% reduction in forward FLOPs compared to CondenseNet and can perform image classification on ARM-Based computing platforms without needing a CUDA enabled GPU support, with outstanding efficiency.
\end{abstract}
\vskip 0.06in

\section{Introduction}
\vskip 0.1in

ARM processors are widely used in electronic devices such as smartphones and tablets as well as in embedded computing platforms such as the NXP BlueBox, Nvidia Jetson and Raspberry Pi for computer vision purposes. ARM is RISC (Reduced Instruction Set Computing) based architecture for computer processors which results in low costs, minimal power consumption, and lower heat generation compared to its competitor: CISC (Complex Instruction Set Computing) architecture based processors such as the Intel x86 processor family. As of 2021, over 180 billion ARM-based chips have been manufactured and shipped by Arm and its partners around the globe which makes it the most popular choice of Instruction Set Architecture (ISA) in the world \cite{10}.

The roots of ARM processors trace back to December 1981 when the first widely successful design, BBC Micro (British Broadcasting Corporation Microcomputer System), was introduced by Acorn Computers \cite{6}. Due to the use of DRAM (Dynamic Random Access Memory) in its design, it outperformed nearly twice as that of Apple II, an 8-bit personal computer, which was world's first successfully mass-produced publicly available computer designed by Steve Wozniak, Steve Jobs and Rod Holt in June 1977 \cite{12}.

Fast forwarding to the 21\textsuperscript{st} century, due to constant advances in computing and VLSI technology, ARM-based chips are found in nearly 60\% of all mobile devices and computing platforms produced today. With processor performance doubling approximately every two years with a focus on parallel computing technologies such as multi-core processors, computer vision researchers can now implement sophisticated neural network algorithms to perform complex computations for OpenCV applications without a requiring a GPU support.

Convolutional Neural Networks (CNN), a class of Deep Neural Networks (DNN) first introduced by Alexey G. Ivakhnenko and V. G. Lapa in 1967 \cite{9}, have been gaining popularity in recent years as researchers focus on creating more advanced intelligent systems. CNNs are popularly used in machine (computer) vision applications such as image classification, image segmentation, object detection, etc. However, implementing a CNN on embedded systems with constrained computational resources for applications such as autonomous cars, robotics and unmanned aerial vehicle (UAV), commonly known as a drone, is a challenging task. In this paper, we present image classification performance results of CondenseNeXt CNN on NXP BlueBox, an ARM-based embedded computing platform for automotive applications.

\section{Related Work}
\vskip 0.02in
Following work has contributed to the research and implementation results presented within this paper:

\vspace{1mm}

\noindent
\textbf{CondenseNeXt:}
An ultra-efficient deep convolutional neural network for embedded systems, introduced by P. Kalgaonkar and M. El-Sharkawy in January 2021 \cite{11} has been utilized to train and evaluate image classification performance on three benchmarking datasets: CIFAR-10, CIFAR-100 and ImageNet.

\section{NXP BlueBox 2.0}
\medskip
The BlueBox 2 family developed and manufactured by NXP Semiconductors N.V, a Dutch-American semiconductor manufacturer with headquarters in Eindhoven, Netherlands and Austin, United States of America, is a Automotive High Performance Compute (AHPC) platform that provides essential performance and reliability for engineers to develop sensor fusion, automated drive and motion planning applications along with functional safety, vision acceleration and automotive interfaces for self-driving (autonomous) vehicles.

NXP BlueBox Gen1 was first introduced in May 2016 at the 2016 NXP FTF Technology Forum held in Austin, Texas, USA. This opened avenues to a host of autonomous and sensor fusion applications. Shortly after, NXP introduced BlueBox Gen2 (BlueBox 2.0), a significant improvement over Gen1, incorporating three new processors: S32V234 ARM-based automotive computer vision processor, LS2084A high performance ARM-based compute processor and S32R274 ASIL-D RADAR microcontroller.
\medskip

\noindent
\textbf{S32V234:}
The S32V234 automotive computer vision processor comprises of a quad core ARM Cortex-A53 CPU running at 1.0 GHz paired with a ARM Cortex-M4 functional safety core which utilizes the ARMv8-A 64-bit instruction set developed by ARM Holdings' Cambridge design centre. It has a 4MB internal SRAM in addition to a 32bit LPDDR3 memory controller for external memory support. It is an on-chip Image Signal Processor (ISP) designed to meet ASIL-B/C automotive safety standards and optimized for obtaining maximum performance per watt efficiency.

\medskip
\noindent
\textbf{LS2084A:}
The LS2084A high performance compute processor comprises of an octa core ARM Cortex-A72 CPU running at 1.8 GHz which utilizes the ARMv8-A 64-bit instruction set developed by ARM Holdings' Austin design centre. It has two 72 bytes DDR4 RAMs running at up to 28.8GB/s memory bandwidth. The LS2 provides software compatibility with next generation LayerScape LX2 family and offers AEC Q100 Grade 3 reliability with 15 years product longevity.

\medskip
\noindent
\textbf{S32R274:}
The S32R274 radar micro-controller comprises of a dual core Freescale PowerPC e200z7 32-bit CPU running at 240 MHz and a dual core Freescale PowerPC e200z4 32-bit CPU running at 120 MHz with an additional checker core. It has a 2 MB Flash and 1.5 MB SRAM for radar application storage, message buffering and radar data stream handling. The S32R processor is optimized for on-chip radar signal processing to maximize performance per watt efficiency. It has been designed by NXP to meet the ASIL-D automotive applications standards.

\begin{figure}
    \centering
    \includegraphics[scale=0.34]{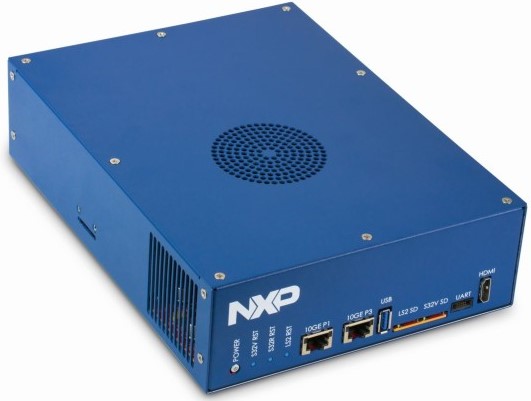}
    \caption{NXP BlueBox 2.0 ARM-based Automotive High Performance Compute (AHPC) embedded development platform. It delivers necessary prerequisites to help develop high-performance computing systems, analyze driving environments, assess risk factors, and then direct the car's behavior. BlueBox 2.0 also supports OpenCV applications using an external camera for real-time image classification object detection and image segmentation.}
    \label{fig:bbphoto}
\end{figure}

\begin{figure}
    \centering
    \includegraphics[scale=0.34]{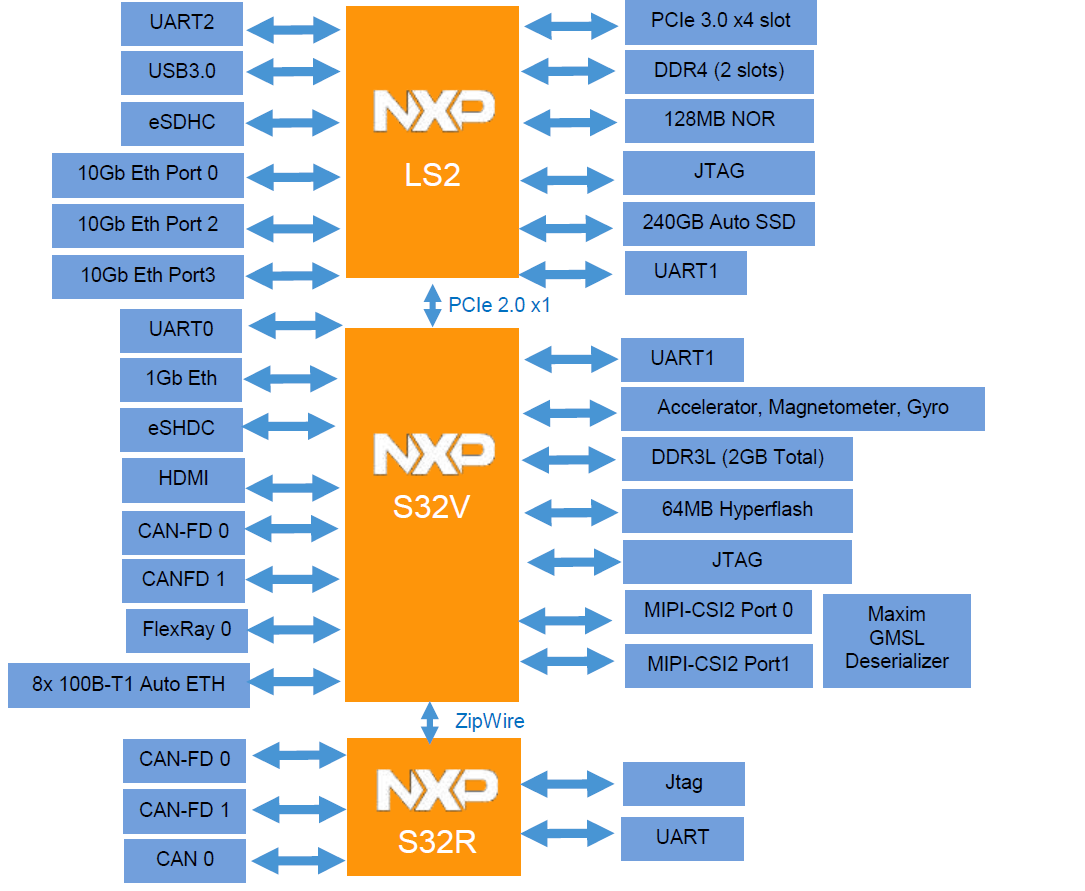}
    \caption{High-Level View of NXP Bluebox 2.0 Gen2 Architecture \cite{13}. S32V processor utilizes two CAN-FD (Flexible Data Rate) with enhanced payload and data rate, PCIe, Ethernet, FlexRay, Zipwire, one SAR-ADCs, four SPI and one SD card connectivity. LS2 processor utilizes two DUART, four I2C, SPIO, GPIO and two USB 3.0 interfaces. S32R processor utilizes JTAG, UART, three FlexCAN and Zipwire to connect to a radar ASIC.}
    \label{fig:my_label2}
\end{figure}

\section{RTMaps Remote Studio Software}

RTMaps (Real-Time Multisensor applications) developed by Intempora is a powerful GUI software that aids in development of applications for advanced driver assistance systems, autonomous driving and robotics. It helps in capturing, processing and viewing data from multiple sensors and offers a multi-modular development and run-time environment for ARM-based computing platforms such as the NXP BlueBox 2.0. This data can also be reviewed and play-backed at a later time for offline development and testing purposes.

RTMaps Remote Studio supports PyTorch, an open-source machine learning library based upon the Torch library, widely used for real-time computer vision (OpenCV) development. Algorithms for OpenCV can be developed using Python scripting language and by the means of block diagrams. It also facilitates the development of algorithms directly on to any supported embedded system without having to connect external user interfacing peripheral devices.

\section{CondenseNeXt}
\smallskip
CondenseNeXt is an ultra-efficient deep convolutional neural network architecture designed for embedded systems introduced by P. Kalgaonkar and M. El-Sharkawy in January  2021. CondenseNeXt refers to the \textit{next} dimension of cardinality. In this section, we describe in detail the architecture of this neural network that has been utilized to train and evaluate image classification performance on three benchmarking datasets: CIFAR-10, CIFAR-100 and ImageNet.

\subsection{Convolution Layers}

One of the main goals of CondenseNeXt is to reduce the amount of computational resources required to train the network from scratch and for real-time inference on embedded systems with limited computational resources. Following state-of-the-art technique has been incorporated into the design of this CNN:
\begin{itemize}
\setlength\itemsep{0.35em}
    \item Depthwise convolution layer: It acts like a filtering layer where convolution to a single input channel is applied separately instead of applying it to all input channels. Assume there is an input data of size $A \times A \times C$ and filters (kernels) $K$ of size $F \times F \times 1$. If there are $C$ number of channels in the input data, the output will be of size $B \times B \times C$. At this point, the spatial dimensions have shrunk. However, the depth $C$ has remained constant and the cost of this operation will be $B^2 \times F^2 \times C$.
    \item Pointwise convolution layer: It acts like a combining layer where a linear combination is carried out for each of these layers. At this stage, a $1 \times 1$ convolution is applied to $C$ number of channels in the input data. Thus, the size of filter for this operation will be $1 \times 1 \times C$ and size of the output will be $B \times B \times D$ for $D$ such filters.
\end{itemize}
\vskip 0.02in
Assume a standard convolutional filter $K$ of size $F \times F \times A \times B$ where $A$ is the number of input channels and $B$ is the number of output channels with an input feature map \textbf{$A$} of size $D_{x} \times D_{x} \times A$ that produces an output feature map \textbf{$Z$} of size $D_{y} \times D_{y}\times B$ can be mathematically represented as follows:

\begin{equation}
\label{eqn_1}
Z_{k,l,n} = \sum_{i,j,m} k_{i,j,m,n} \cdot A_{k+i-1,l+j-1,m}
\end{equation}

In case of a depthwise separable convolution, (\ref{eqn_1}) is factorized into two stages: the first stage applies a $3 \times 3$ depthwise convolution $\hat{K}$ with one filter for every input channel:

\begin{equation}
\label{eqn_2}
\hat{Z}_{k,l,m} = \sum_{i,j} \hat{K}_{i,j,m} \cdot A_{k+i-1,l+j-1,m}
\end{equation}

Consequently, in the second stage, a $1 \times 1$ pointwise convolution $\tilde{K}$ is applied to carry out linear combination and combine the outputs of depthwise convolution from previous stage as follows:

\begin{equation}
\label{eqn_3}
Z_{k,l,n} = \sum_{m} \tilde{K}_{m,n} \cdot \hat{Z}_{k-1,l-1,m}
\end{equation}

\begin{figure}[t]
\tikzset{every picture/.style={line width=0.75pt}} 

\begin{tikzpicture}[scale=0.51,x=0.75pt,y=0.75pt,yscale=-1,xscale=1]

\draw  [fill={rgb, 255:red, 65; green, 117; blue, 5 }  ,fill opacity=1 ] (202.96,182.02) -- (229.1,197.97) -- (220.49,203.32) -- (194.34,187.37) -- (194.34,187.37) -- cycle ;
\draw  [fill={rgb, 255:red, 65; green, 117; blue, 5 }  ,fill opacity=1 ] (194.34,213.14) -- (194.34,187.38) -- (220.49,203.32) -- (220.49,229.08) -- cycle ;
\draw [fill={rgb, 255:red, 65; green, 117; blue, 5 }  ,fill opacity=1 ]   (211.9,197.68) -- (211.98,223.88) ;
\draw [fill={rgb, 255:red, 65; green, 117; blue, 5 }  ,fill opacity=1 ]   (203.65,192.93) -- (203.47,218.49) ;
\draw  [fill={rgb, 255:red, 65; green, 117; blue, 5 }  ,fill opacity=1 ] (229.1,223.55) -- (220.49,228.73) -- (220.49,203.15) -- (229.1,197.97) -- cycle ;
\draw [fill={rgb, 255:red, 65; green, 117; blue, 5 }  ,fill opacity=1 ]   (220.49,212.28) -- (229.1,207.1) ;
\draw [fill={rgb, 255:red, 65; green, 117; blue, 5 }  ,fill opacity=1 ]   (220.49,221.27) -- (229.1,216.08) ;
\draw [fill={rgb, 255:red, 65; green, 117; blue, 5 }  ,fill opacity=1 ]   (211.9,197.68) -- (220.52,192.49) ;
\draw [fill={rgb, 255:red, 65; green, 117; blue, 5 }  ,fill opacity=1 ]   (203,193.44) -- (211.62,188.25) ;
\draw [fill={rgb, 255:red, 65; green, 117; blue, 5 }  ,fill opacity=1 ]   (194.09,187.87) -- (202.71,182.52) ;
\draw [fill={rgb, 255:red, 65; green, 117; blue, 5 }  ,fill opacity=1 ]   (194.34,205.32) -- (220.49,221.27) ;
\draw [fill={rgb, 255:red, 65; green, 117; blue, 5 }  ,fill opacity=1 ]   (194.34,196.34) -- (220.49,212.28) ;
\draw [fill={rgb, 255:red, 65; green, 117; blue, 5 }  ,fill opacity=1 ]   (202.71,182.52) -- (228.85,198.46) ;

\draw  [fill={rgb, 255:red, 208; green, 2; blue, 27 }  ,fill opacity=1 ] (189.96,190.02) -- (216.1,205.97) -- (207.49,211.32) -- (181.34,195.37) -- (181.34,195.37) -- cycle ;
\draw  [fill={rgb, 255:red, 208; green, 2; blue, 27 }  ,fill opacity=1 ] (181.34,221.14) -- (181.34,195.38) -- (207.49,211.32) -- (207.49,237.08) -- cycle ;
\draw [fill={rgb, 255:red, 208; green, 2; blue, 27 }  ,fill opacity=1 ]   (198.9,205.68) -- (198.98,231.88) ;
\draw [fill={rgb, 255:red, 208; green, 2; blue, 27 }  ,fill opacity=1 ]   (190.65,200.93) -- (190.47,226.49) ;
\draw  [fill={rgb, 255:red, 208; green, 2; blue, 27 }  ,fill opacity=1 ] (216.1,231.55) -- (207.49,236.73) -- (207.49,211.15) -- (216.1,205.97) -- cycle ;
\draw [fill={rgb, 255:red, 208; green, 2; blue, 27 }  ,fill opacity=1 ]   (207.49,220.28) -- (216.1,215.1) ;
\draw [fill={rgb, 255:red, 208; green, 2; blue, 27 }  ,fill opacity=1 ]   (207.49,229.27) -- (216.1,224.08) ;
\draw [fill={rgb, 255:red, 208; green, 2; blue, 27 }  ,fill opacity=1 ]   (198.9,205.68) -- (207.52,200.49) ;
\draw [fill={rgb, 255:red, 208; green, 2; blue, 27 }  ,fill opacity=1 ]   (190,201.44) -- (198.62,196.25) ;
\draw [fill={rgb, 255:red, 208; green, 2; blue, 27 }  ,fill opacity=1 ]   (181.09,195.87) -- (189.71,190.52) ;
\draw [fill={rgb, 255:red, 208; green, 2; blue, 27 }  ,fill opacity=1 ]   (181.34,213.32) -- (207.49,229.27) ;
\draw [fill={rgb, 255:red, 208; green, 2; blue, 27 }  ,fill opacity=1 ]   (181.34,204.34) -- (207.49,220.28) ;
\draw [fill={rgb, 255:red, 208; green, 2; blue, 27 }  ,fill opacity=1 ]   (189.71,190.52) -- (215.85,206.46) ;

\draw  [fill={rgb, 255:red, 80; green, 227; blue, 194 }  ,fill opacity=0.69 ] (628.51,161.8) -- (666.53,194.02) -- (581.93,245.12) -- (543.9,212.91) -- (628.51,161.8) -- cycle ;
\draw  [fill={rgb, 255:red, 80; green, 227; blue, 194 }  ,fill opacity=0.64 ] (581.93,245.12) -- (581.93,284.92) -- (543.9,252.69) -- (543.9,212.9) -- cycle ;
\draw  [fill={rgb, 255:red, 80; green, 227; blue, 194 }  ,fill opacity=0.64 ] (666.53,194.02) -- (666.53,233.79) -- (581.93,284.91) -- (581.93,245.14) -- cycle ;
\draw [fill={rgb, 255:red, 80; green, 227; blue, 194 }  ,fill opacity=0.64 ]   (543.9,221.06) -- (581.93,253.28) ;
\draw [fill={rgb, 255:red, 80; green, 227; blue, 194 }  ,fill opacity=0.64 ]   (543.9,229.02) -- (581.93,261.23) ;
\draw [fill={rgb, 255:red, 80; green, 227; blue, 194 }  ,fill opacity=0.64 ]   (543.9,236.58) -- (581.93,268.8) ;
\draw [fill={rgb, 255:red, 80; green, 227; blue, 194 }  ,fill opacity=0.64 ]   (543.9,244.54) -- (581.93,276.76) ;
\draw [fill={rgb, 255:red, 80; green, 227; blue, 194 }  ,fill opacity=0.64 ]   (559.11,225.83) -- (559.11,265.62) ;
\draw [fill={rgb, 255:red, 80; green, 227; blue, 194 }  ,fill opacity=0.64 ]   (551.51,219.47) -- (551.51,259.25) ;
\draw [fill={rgb, 255:red, 80; green, 227; blue, 194 }  ,fill opacity=0.64 ]   (566.72,232.2) -- (566.72,271.99) ;
\draw [fill={rgb, 255:red, 80; green, 227; blue, 194 }  ,fill opacity=0.64 ]   (574.32,238.56) -- (574.32,278.35) ;
\draw [fill={rgb, 255:red, 80; green, 227; blue, 194 }  ,fill opacity=0.64 ]   (597.14,236.18) -- (597.14,275.96) ;
\draw [fill={rgb, 255:red, 80; green, 227; blue, 194 }  ,fill opacity=0.64 ]   (589.53,240.95) -- (589.53,280.74) ;
\draw [fill={rgb, 255:red, 80; green, 227; blue, 194 }  ,fill opacity=0.64 ]   (612.35,226.63) -- (612.35,266.41) ;
\draw [fill={rgb, 255:red, 80; green, 227; blue, 194 }  ,fill opacity=0.64 ]   (604.75,231.4) -- (604.75,271.19) ;
\draw [fill={rgb, 255:red, 80; green, 227; blue, 194 }  ,fill opacity=0.64 ]   (581.93,252.89) -- (612.35,234.59) ;
\draw [fill={rgb, 255:red, 80; green, 227; blue, 194 }  ,fill opacity=0.64 ]   (581.93,260.84) -- (612.35,242.54) ;
\draw [fill={rgb, 255:red, 80; green, 227; blue, 194 }  ,fill opacity=0.64 ]   (581.93,268.8) -- (612.35,250.5) ;
\draw [fill={rgb, 255:red, 80; green, 227; blue, 194 }  ,fill opacity=0.64 ]   (581.93,276.76) -- (612.35,258.46) ;
\draw [fill={rgb, 255:red, 80; green, 227; blue, 194 }  ,fill opacity=0.64 ][line width=2.25]  [dash pattern={on 2.53pt off 3.02pt}]  (624.23,239.47) -- (651.61,221.96) ;
\draw [fill={rgb, 255:red, 80; green, 227; blue, 194 }  ,fill opacity=0.64 ]   (628.51,161.8) -- (543.9,212.91) ;
\draw [fill={rgb, 255:red, 80; green, 227; blue, 194 }  ,fill opacity=0.64 ]   (628.51,161.8) -- (666.53,194.02) ;
\draw [fill={rgb, 255:red, 80; green, 227; blue, 194 }  ,fill opacity=0.64 ][line width=2.25]  [dash pattern={on 2.53pt off 3.02pt}]  (606.84,203.06) -- (634.22,185.55) ;
\draw [fill={rgb, 255:red, 80; green, 227; blue, 194 }  ,fill opacity=0.64 ]   (551.51,208.73) -- (589.53,240.95) ;
\draw [fill={rgb, 255:red, 80; green, 227; blue, 194 }  ,fill opacity=0.64 ]   (574.32,194.41) -- (612.35,226.63) ;
\draw [fill={rgb, 255:red, 80; green, 227; blue, 194 }  ,fill opacity=0.64 ]   (566.72,199.18) -- (604.75,231.4) ;
\draw [fill={rgb, 255:red, 80; green, 227; blue, 194 }  ,fill opacity=0.64 ]   (559.11,203.96) -- (597.14,236.18) ;
\draw [fill={rgb, 255:red, 80; green, 227; blue, 194 }  ,fill opacity=0.64 ]   (551.51,219.47) -- (581.93,201.16) ;
\draw [fill={rgb, 255:red, 80; green, 227; blue, 194 }  ,fill opacity=0.64 ]   (559.11,225.83) -- (589.53,207.53) ;
\draw [fill={rgb, 255:red, 80; green, 227; blue, 194 }  ,fill opacity=0.64 ]   (566.72,232.2) -- (597.14,213.9) ;
\draw [fill={rgb, 255:red, 80; green, 227; blue, 194 }  ,fill opacity=0.64 ]   (574.32,238.56) -- (604.75,220.26) ;

\draw  [fill={rgb, 255:red, 245; green, 166; blue, 35 }  ,fill opacity=1 ] (429.93,223.72) -- (429.93,223.72) -- (437.33,228) -- (467.33,210.68) -- (459.93,206.4) -- cycle ;
\draw  [fill={rgb, 255:red, 245; green, 166; blue, 35 }  ,fill opacity=1 ] (467.33,210.68) -- (467.33,220.68) -- (437.33,238) -- (437.33,228) -- cycle ;
\draw  [fill={rgb, 255:red, 245; green, 166; blue, 35 }  ,fill opacity=1 ] (437.33,228) -- (437.33,238.26) -- (429.93,233.98) -- (429.93,223.72) -- cycle ;
\draw [fill={rgb, 255:red, 245; green, 166; blue, 35 }  ,fill opacity=1 ]   (429.93,223.72) -- (459.93,206.4) ;
\draw [fill={rgb, 255:red, 245; green, 166; blue, 35 }  ,fill opacity=1 ]   (457.33,216.9) -- (457.33,227.16) ;
\draw [fill={rgb, 255:red, 245; green, 166; blue, 35 }  ,fill opacity=1 ]   (447.11,222.56) -- (447.11,232.82) ;
\draw [fill={rgb, 255:red, 245; green, 166; blue, 35 }  ,fill opacity=1 ]   (449.93,212.63) -- (457.33,216.9) ;
\draw [fill={rgb, 255:red, 245; green, 166; blue, 35 }  ,fill opacity=1 ]   (439.71,218.29) -- (447.11,222.56) ;

\draw  [fill={rgb, 255:red, 65; green, 117; blue, 5 }  ,fill opacity=1 ] (319.84,172.21) -- (360.4,199.09) -- (352.29,204.5) -- (352.29,204.5) -- (311.72,177.62) -- cycle ;
\draw  [fill={rgb, 255:red, 65; green, 117; blue, 5 }  ,fill opacity=1 ] (311.72,224.43) -- (311.72,177.62) -- (352.29,204.47) -- (352.29,251.28) -- cycle ;
\draw  [fill={rgb, 255:red, 65; green, 117; blue, 5 }  ,fill opacity=1 ] (360.4,245.88) -- (352.29,251.28) -- (352.29,204.5) -- (360.4,199.09) -- cycle ;
\draw [fill={rgb, 255:red, 65; green, 117; blue, 5 }  ,fill opacity=1 ]   (352.29,224.94) -- (360.4,219.54) ;
\draw [fill={rgb, 255:red, 65; green, 117; blue, 5 }  ,fill opacity=1 ]   (352.29,234.31) -- (360.4,228.9) ;
\draw [fill={rgb, 255:red, 65; green, 117; blue, 5 }  ,fill opacity=1 ]   (352.29,215.58) -- (360.4,210.18) ;
\draw [fill={rgb, 255:red, 65; green, 117; blue, 5 }  ,fill opacity=1 ]   (319.84,183.05) -- (327.95,177.64) ;
\draw [fill={rgb, 255:red, 65; green, 117; blue, 5 }  ,fill opacity=1 ]   (327.95,188.55) -- (336.06,183.15) ;
\draw [fill={rgb, 255:red, 65; green, 117; blue, 5 }  ,fill opacity=1 ]   (336.06,193.51) -- (344.17,188.1) ;
\draw [fill={rgb, 255:red, 65; green, 117; blue, 5 }  ,fill opacity=1 ]   (344.17,198.85) -- (352.29,193.45) ;
\draw [fill={rgb, 255:red, 65; green, 117; blue, 5 }  ,fill opacity=1 ]   (344.17,198.85) -- (344.17,246.11) ;
\draw [fill={rgb, 255:red, 65; green, 117; blue, 5 }  ,fill opacity=1 ]   (336.06,193.51) -- (336.06,240.76) ;
\draw [fill={rgb, 255:red, 65; green, 117; blue, 5 }  ,fill opacity=1 ]   (327.95,188.55) -- (327.95,235.81) ;
\draw [fill={rgb, 255:red, 65; green, 117; blue, 5 }  ,fill opacity=1 ]   (319.84,183.05) -- (319.84,230.3) ;
\draw [fill={rgb, 255:red, 65; green, 117; blue, 5 }  ,fill opacity=1 ]   (311.72,188.73) -- (352.29,215.58) ;
\draw [fill={rgb, 255:red, 65; green, 117; blue, 5 }  ,fill opacity=1 ]   (311.72,215.98) -- (352.29,242.84) ;
\draw [fill={rgb, 255:red, 65; green, 117; blue, 5 }  ,fill opacity=1 ]   (311.72,207.45) -- (352.29,234.31) ;
\draw [fill={rgb, 255:red, 65; green, 117; blue, 5 }  ,fill opacity=1 ]   (311.72,198.09) -- (352.29,224.94) ;
\draw [fill={rgb, 255:red, 65; green, 117; blue, 5 }  ,fill opacity=1 ]   (352.29,242.84) -- (360.4,237.43) ;

\draw  [fill={rgb, 255:red, 208; green, 2; blue, 27 }  ,fill opacity=1 ] (311.72,177.83) -- (352.29,204.71) -- (344.17,210.12) -- (344.17,210.12) -- (303.61,183.23) -- cycle ;
\draw  [fill={rgb, 255:red, 208; green, 2; blue, 27 }  ,fill opacity=1 ] (303.61,230.05) -- (303.61,183.24) -- (344.17,210.09) -- (344.17,256.89) -- cycle ;
\draw  [fill={rgb, 255:red, 208; green, 2; blue, 27 }  ,fill opacity=1 ] (352.29,251.49) -- (344.17,256.9) -- (344.17,210.12) -- (352.29,204.71) -- cycle ;
\draw [fill={rgb, 255:red, 208; green, 2; blue, 27 }  ,fill opacity=1 ]   (344.17,230.56) -- (352.29,225.15) ;
\draw [fill={rgb, 255:red, 208; green, 2; blue, 27 }  ,fill opacity=1 ]   (344.17,239.92) -- (352.29,234.52) ;
\draw [fill={rgb, 255:red, 208; green, 2; blue, 27 }  ,fill opacity=1 ]   (344.17,221.2) -- (352.29,215.79) ;
\draw [fill={rgb, 255:red, 208; green, 2; blue, 27 }  ,fill opacity=1 ]   (311.72,188.67) -- (319.84,183.26) ;
\draw [fill={rgb, 255:red, 208; green, 2; blue, 27 }  ,fill opacity=1 ]   (319.84,194.17) -- (327.95,188.77) ;
\draw [fill={rgb, 255:red, 208; green, 2; blue, 27 }  ,fill opacity=1 ]   (327.95,199.12) -- (336.06,193.72) ;
\draw [fill={rgb, 255:red, 208; green, 2; blue, 27 }  ,fill opacity=1 ]   (336.06,204.47) -- (344.17,199.06) ;
\draw    (336.06,204.47) -- (336.06,251.72) ;
\draw    (327.95,199.12) -- (327.95,246.38) ;
\draw    (319.84,194.17) -- (319.84,241.43) ;
\draw    (311.72,188.67) -- (311.72,235.92) ;
\draw    (303.61,194.35) -- (344.17,221.2) ;
\draw    (303.61,221.6) -- (344.17,248.45) ;
\draw    (303.61,213.07) -- (344.17,239.92) ;
\draw    (303.61,203.71) -- (344.17,230.56) ;
\draw [fill={rgb, 255:red, 208; green, 2; blue, 27 }  ,fill opacity=1 ]   (344.17,248.45) -- (352.29,243.05) ;

\draw  [fill={rgb, 255:red, 245; green, 166; blue, 35 }  ,fill opacity=1 ] (303.61,183.45) -- (344.17,210.33) -- (336.06,215.73) -- (336.06,215.73) -- (295.5,188.85) -- cycle ;
\draw  [fill={rgb, 255:red, 245; green, 166; blue, 35 }  ,fill opacity=1 ] (295.5,235.66) -- (295.5,188.86) -- (336.06,215.7) -- (336.06,262.51) -- cycle ;
\draw  [fill={rgb, 255:red, 245; green, 166; blue, 35 }  ,fill opacity=1 ] (344.17,257.11) -- (336.06,262.52) -- (336.06,215.73) -- (344.17,210.33) -- cycle ;
\draw [fill={rgb, 255:red, 245; green, 166; blue, 35 }  ,fill opacity=1 ]   (336.06,236.18) -- (344.17,230.77) ;
\draw [fill={rgb, 255:red, 245; green, 166; blue, 35 }  ,fill opacity=1 ]   (336.06,245.54) -- (344.17,240.13) ;
\draw [fill={rgb, 255:red, 245; green, 166; blue, 35 }  ,fill opacity=1 ]   (336.06,226.82) -- (344.17,221.41) ;
\draw [fill={rgb, 255:red, 245; green, 166; blue, 35 }  ,fill opacity=1 ]   (303.61,194.28) -- (311.72,188.88) ;
\draw [fill={rgb, 255:red, 245; green, 166; blue, 35 }  ,fill opacity=1 ]   (311.72,199.79) -- (319.84,194.38) ;
\draw [fill={rgb, 255:red, 245; green, 166; blue, 35 }  ,fill opacity=1 ]   (319.84,204.74) -- (327.95,199.34) ;
\draw [fill={rgb, 255:red, 245; green, 166; blue, 35 }  ,fill opacity=1 ]   (327.95,210.09) -- (336.06,204.68) ;
\draw [fill={rgb, 255:red, 245; green, 166; blue, 35 }  ,fill opacity=1 ]   (327.95,210.09) -- (327.95,257.34) ;
\draw [fill={rgb, 255:red, 245; green, 166; blue, 35 }  ,fill opacity=1 ]   (319.84,204.74) -- (319.84,252) ;
\draw [fill={rgb, 255:red, 245; green, 166; blue, 35 }  ,fill opacity=1 ]   (311.72,199.79) -- (311.72,247.04) ;
\draw [fill={rgb, 255:red, 245; green, 166; blue, 35 }  ,fill opacity=1 ]   (303.61,194.28) -- (303.61,241.54) ;
\draw [fill={rgb, 255:red, 245; green, 166; blue, 35 }  ,fill opacity=1 ]   (295.5,199.96) -- (336.06,226.82) ;
\draw [fill={rgb, 255:red, 245; green, 166; blue, 35 }  ,fill opacity=1 ]   (295.5,227.22) -- (336.06,254.07) ;
\draw [fill={rgb, 255:red, 245; green, 166; blue, 35 }  ,fill opacity=1 ]   (295.5,218.69) -- (336.06,245.54) ;
\draw [fill={rgb, 255:red, 245; green, 166; blue, 35 }  ,fill opacity=1 ]   (295.5,209.32) -- (336.06,236.18) ;
\draw [fill={rgb, 255:red, 245; green, 166; blue, 35 }  ,fill opacity=1 ]   (336.06,254.07) -- (344.17,248.66) ;

\draw  [color={rgb, 255:red, 245; green, 166; blue, 35 }  ,draw opacity=1 ][dash pattern={on 4.5pt off 4.5pt}] (151.81,218.81) .. controls (151.81,193.39) and (172.57,172.79) .. (198.17,172.79) .. controls (223.78,172.79) and (244.53,193.39) .. (244.53,218.81) .. controls (244.53,244.22) and (223.78,264.83) .. (198.17,264.83) .. controls (172.57,264.83) and (151.81,244.22) .. (151.81,218.81) -- cycle ;
\draw  [fill={rgb, 255:red, 245; green, 166; blue, 35 }  ,fill opacity=1 ] (176.96,198.27) -- (203.1,214.22) -- (194.49,219.57) -- (168.34,203.62) -- (168.34,203.62) -- cycle ;
\draw  [fill={rgb, 255:red, 245; green, 166; blue, 35 }  ,fill opacity=1 ] (168.34,229.39) -- (168.34,203.63) -- (194.49,219.57) -- (194.49,245.33) -- cycle ;
\draw [fill={rgb, 255:red, 245; green, 166; blue, 35 }  ,fill opacity=1 ]   (185.9,213.93) -- (185.98,240.13) ;
\draw [fill={rgb, 255:red, 245; green, 166; blue, 35 }  ,fill opacity=1 ]   (177.65,209.18) -- (177.47,234.74) ;
\draw  [fill={rgb, 255:red, 245; green, 166; blue, 35 }  ,fill opacity=1 ] (203.1,239.8) -- (194.49,244.98) -- (194.49,219.4) -- (203.1,214.22) -- cycle ;
\draw    (194.49,228.53) -- (203.1,223.35) ;
\draw    (194.49,237.52) -- (203.1,232.33) ;
\draw    (185.9,213.93) -- (194.52,208.74) ;
\draw    (177,209.69) -- (185.62,204.5) ;
\draw    (168.09,204.12) -- (176.71,198.77) ;
\draw    (168.34,221.57) -- (194.49,237.52) ;
\draw    (168.34,212.59) -- (194.49,228.53) ;
\draw    (176.71,198.77) -- (202.85,214.71) ;

\draw  [color={rgb, 255:red, 245; green, 166; blue, 35 }  ,draw opacity=1 ][dash pattern={on 4.5pt off 4.5pt}] (405.15,221.75) .. controls (405.15,197.47) and (424.83,177.79) .. (449.11,177.79) .. controls (473.39,177.79) and (493.07,197.47) .. (493.07,221.75) .. controls (493.07,246.03) and (473.39,265.71) .. (449.11,265.71) .. controls (424.83,265.71) and (405.15,246.03) .. (405.15,221.75) -- cycle ;
\draw  [fill={rgb, 255:red, 74; green, 144; blue, 226 }  ,fill opacity=1 ] (70.34,163.57) -- (101.6,181.46) -- (80.31,193.5) -- (80.31,193.5) -- (49.05,175.69) -- cycle ;
\draw  [fill={rgb, 255:red, 74; green, 144; blue, 226 }  ,fill opacity=1 ] (24.65,228.09) -- (24.65,161.8) -- (80.31,193.5) -- (80.31,259.78) -- cycle ;
\draw [fill={rgb, 255:red, 74; green, 144; blue, 226 }  ,fill opacity=1 ]   (80.31,252.57) -- (24.65,220.87) ;
\draw [fill={rgb, 255:red, 74; green, 144; blue, 226 }  ,fill opacity=1 ]   (80.31,242.72) -- (24.65,211.02) ;
\draw [fill={rgb, 255:red, 74; green, 144; blue, 226 }  ,fill opacity=1 ]   (80.31,203.34) -- (24.65,171.65) ;
\draw [fill={rgb, 255:red, 74; green, 144; blue, 226 }  ,fill opacity=1 ]   (80.31,213.19) -- (24.65,181.49) ;
\draw [fill={rgb, 255:red, 74; green, 144; blue, 226 }  ,fill opacity=1 ]   (80.31,223.03) -- (24.65,191.33) ;
\draw [fill={rgb, 255:red, 74; green, 144; blue, 226 }  ,fill opacity=1 ]   (80.31,232.88) -- (24.65,201.18) ;

\draw [fill={rgb, 255:red, 74; green, 144; blue, 226 }  ,fill opacity=1 ]   (73.45,189.59) -- (73.45,255.87) ;
\draw [fill={rgb, 255:red, 74; green, 144; blue, 226 }  ,fill opacity=1 ]   (32.79,166.43) -- (32.79,232.72) ;
\draw [fill={rgb, 255:red, 74; green, 144; blue, 226 }  ,fill opacity=1 ]   (41.54,171.42) -- (41.54,237.7) ;
\draw [fill={rgb, 255:red, 74; green, 144; blue, 226 }  ,fill opacity=1 ]   (49.05,175.69) -- (49.05,241.98) ;
\draw [fill={rgb, 255:red, 74; green, 144; blue, 226 }  ,fill opacity=1 ]   (57.18,180.33) -- (57.18,246.61) ;
\draw [fill={rgb, 255:red, 74; green, 144; blue, 226 }  ,fill opacity=1 ]   (65.31,184.96) -- (65.31,251.24) ;

\draw  [fill={rgb, 255:red, 74; green, 144; blue, 226 }  ,fill opacity=1 ] (101.6,247.73) -- (80.31,259.85) -- (80.31,193.59) -- (101.6,181.46) -- cycle ;
\draw [fill={rgb, 255:red, 74; green, 144; blue, 226 }  ,fill opacity=1 ]   (87.41,189.55) -- (87.41,255.81) ;
\draw [fill={rgb, 255:red, 74; green, 144; blue, 226 }  ,fill opacity=1 ]   (94.5,185.5) -- (94.5,251.77) ;
\draw [fill={rgb, 255:red, 74; green, 144; blue, 226 }  ,fill opacity=1 ]   (80.31,232.88) -- (101.6,220.76) ;
\draw [fill={rgb, 255:red, 74; green, 144; blue, 226 }  ,fill opacity=1 ]   (80.31,242.72) -- (101.6,230.6) ;
\draw [fill={rgb, 255:red, 74; green, 144; blue, 226 }  ,fill opacity=1 ]   (80.31,223.03) -- (101.6,210.91) ;
\draw [fill={rgb, 255:red, 74; green, 144; blue, 226 }  ,fill opacity=1 ]   (80.31,252.57) -- (101.6,240.44) ;
\draw [fill={rgb, 255:red, 74; green, 144; blue, 226 }  ,fill opacity=1 ]   (80.31,213.19) -- (101.6,201.07) ;
\draw [fill={rgb, 255:red, 74; green, 144; blue, 226 }  ,fill opacity=1 ]   (80.31,203.34) -- (101.6,191.22) ;

\draw    (24.65,161.8) -- (80.31,193.5) ;
\draw [fill={rgb, 255:red, 74; green, 144; blue, 226 }  ,fill opacity=1 ]   (45.94,149.77) -- (101.6,181.46) ;
\draw [fill={rgb, 255:red, 74; green, 144; blue, 226 }  ,fill opacity=1 ]   (24.65,161.8) -- (45.94,149.68) ;
\draw [fill={rgb, 255:red, 74; green, 144; blue, 226 }  ,fill opacity=1 ]   (31.75,157.85) -- (87.41,189.55) ;
\draw [fill={rgb, 255:red, 74; green, 144; blue, 226 }  ,fill opacity=1 ]   (38.84,153.81) -- (94.5,185.5) ;
\draw [fill={rgb, 255:red, 74; green, 144; blue, 226 }  ,fill opacity=1 ]   (41.54,171.42) -- (62.83,159.3) ;
\draw [fill={rgb, 255:red, 74; green, 144; blue, 226 }  ,fill opacity=1 ]   (49.05,175.69) -- (70.34,163.57) ;
\draw [fill={rgb, 255:red, 74; green, 144; blue, 226 }  ,fill opacity=1 ]   (32.79,166.43) -- (54.07,154.31) ;

\draw [fill={rgb, 255:red, 74; green, 144; blue, 226 }  ,fill opacity=1 ]   (57.18,180.33) -- (78.47,168.2) ;
\draw [fill={rgb, 255:red, 74; green, 144; blue, 226 }  ,fill opacity=1 ]   (65.31,184.96) -- (86.6,172.83) ;

\draw    (80.31,193.59) -- (101.6,181.46) ;
\draw    (73.45,189.59) -- (94.73,177.47) ;
\draw  [fill={rgb, 255:red, 65; green, 117; blue, 5 }  ,fill opacity=0.89 ] (45.94,149.77) -- (70.34,163.57) -- (63.34,167.4) -- (63.34,167.4) -- (38.84,153.81) -- cycle ;
\draw  [fill={rgb, 255:red, 208; green, 2; blue, 27 }  ,fill opacity=0.85 ] (38.84,153.81) -- (63.34,167.4) -- (56.15,171.36) -- (31.75,157.85) -- (31.75,157.85) -- cycle ;
\draw  [fill={rgb, 255:red, 245; green, 166; blue, 35 }  ,fill opacity=0.8 ] (31.75,157.85) -- (56.15,171.36) -- (49.05,175.69) -- (24.65,161.8) -- (24.65,161.8) -- cycle ;
\draw  [fill={rgb, 255:red, 245; green, 166; blue, 35 }  ,fill opacity=0.81 ] (24.65,161.8) -- (49.05,175.69) -- (48.97,205.01) -- (24.65,191.33) -- (24.65,191.33) -- cycle ;

\draw [color={rgb, 255:red, 245; green, 166; blue, 35 }  ,draw opacity=1 ][line width=2.25]    (118.48,218.14) -- (151.81,218.81) ;
\draw [color={rgb, 255:red, 245; green, 166; blue, 35 }  ,draw opacity=1 ][line width=2.25]    (371.81,221.08) -- (405.15,221.75) ;
\draw [color={rgb, 255:red, 245; green, 166; blue, 35 }  ,draw opacity=1 ][line width=2.25]    (493.07,221.75) -- (521.4,222.31) ;
\draw [shift={(526.4,222.41)}, rotate = 181.15] [fill={rgb, 255:red, 245; green, 166; blue, 35 }  ,fill opacity=1 ][line width=0.08]  [draw opacity=0] (10,-4.8) -- (0,0) -- (10,4.8) -- cycle    ;
\draw [color={rgb, 255:red, 245; green, 166; blue, 35 }  ,draw opacity=1 ][line width=2.25]    (244.53,218.81) -- (272.87,219.37) ;
\draw [shift={(277.87,219.47)}, rotate = 181.15] [fill={rgb, 255:red, 245; green, 166; blue, 35 }  ,fill opacity=1 ][line width=0.08]  [draw opacity=0] (10,-4.8) -- (0,0) -- (10,4.8) -- cycle    ;

\draw (197,244) node [anchor=north west][inner sep=0.75pt]   [align=left] {{\scriptsize $1$}};
\draw (222,228) node [anchor=north west][inner sep=0.75pt]   [align=left] {{\scriptsize $1$}};
\draw (209,235) node [anchor=north west][inner sep=0.75pt]   [align=left] {{\scriptsize $1$}};
\draw (90,253) node [anchor=north west][inner sep=0.75pt]   [align=left] {{\scriptsize 3}};
\draw (171.72,237) node [anchor=north west][inner sep=0.75pt]   [align=left] {{\scriptsize 3}};
\draw (154,213) node [anchor=north west][inner sep=0.75pt]   [align=left] {{\scriptsize 3}};
\draw (10,187) node [anchor=north west][inner sep=0.75pt]   [align=left] {{\scriptsize 7}};
\draw (43.54,244) node [anchor=north west][inner sep=0.75pt]   [align=left] {{\scriptsize 7}};
\draw (552,266) node [anchor=north west][inner sep=0.75pt]   [align=left] {{\scriptsize 5}};
\draw (530,226.5) node [anchor=north west][inner sep=0.75pt]   [align=left] {{\scriptsize 5}};
\draw (620.57,260) node [anchor=north west][inner sep=0.75pt]   [align=left] {{\scriptsize 128}};
\draw (346.17,253) node [anchor=north west][inner sep=0.75pt]   [align=left] {{\scriptsize 3}};
\draw (280,207) node [anchor=north west][inner sep=0.75pt]   [align=left] {{\scriptsize 5}};
\draw (305.61,247) node [anchor=north west][inner sep=0.75pt]   [align=left] {{\scriptsize 5}};
\draw (451.11,229) node [anchor=north west][inner sep=0.75pt]   [align=left] {{\scriptsize 3}};
\draw (415,220) node [anchor=north west][inner sep=0.75pt]   [align=left] {{\scriptsize $1$}};
\draw (424,240) node [anchor=north west][inner sep=0.75pt]   [align=left] {{\scriptsize $1$}};
\draw (85.33,124) node [anchor=north west][inner sep=0.75pt]   [align=left] {\scriptsize Depthwise Convolution Operation};
\draw (380,124) node [anchor=north west][inner sep=0.75pt]   [align=left] {\scriptsize Pointwise Convolution Operation};
\draw (429,156) node [anchor=north west][inner sep=0.75pt]   [align=left] {{\scriptsize x 128}};
\end{tikzpicture}
    \caption{A 3D illustration of the overall process of depthwise separable convolution. An image is transformed 128 times whereas an image is transformed by depthwise separable convolution only once and then this transformed image is stretched to 128 channels which allows the neural network to process more data while consuming fewer FLOPs (Floating Point Operations).}
    \label{fig}
\end{figure}
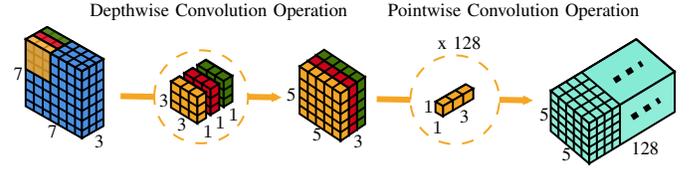

This methodology splits a kernel into two discrete filters for filtering and combining stages as shown in Figure 3 above, which results in reduction of computational resources required to train the network from scratch as well for real-time inference.

A widely used model compression technique is also implemented into the design of this CNN where a further significant impact, both in computational efficiency at training time and on the final trained model size is seen.

\subsection{Model Compression}

A widely popular model compression technique called Group-wise Pruning is implemented to make CondenseNeXt neural network computationally more efficient by discarding redundant elements without influencing the overall performance of the network.
\smallskip
\newline
\textbf{Group-wise Pruning:}
The purpose of group-wise pruning is to remove trivial filters for every group $g$ during the training process which is based on the $L_{1}$-Normalization of $A^{g_{ij}}$ where for every group $g$, $a$ is the input and $z$ is the output. A pruning hyper-parameter $p$ is established and set to $4$ which allows the network to decide the number of filters to remove before the first stage of depthwise separable convolution. A class balanced focal loss function \cite{14} is also added to assist and ease the effect of this pruning process.

Consider a group convolution comprised of $G$ groups of size $F \times F \times C_{A} \times C_{B}$ where $C_{A}=$\( \frac{A}{G} \) and $C_{B}=$\( \frac{B}{G} \). The total number of trivial filters that will be pruned before the first stage of depthwise separable convolution is mathematically represented as follows:
\begin{equation}
\label{eqn_4}
G \cdot C_{x}=A \cdot C-p \cdot A
\end{equation}

\noindent
\textbf{Cardinality:}
A new dimension to the network called \textit{Cardinality} denoted by ${C}$ is incorporated into the design of CondenseNeXt neural network in addition to the existing width and depth dimensions so that loss in accuracy during the pruning process is reduced. Experiments prove that increasing cardinality is a more efficacious way of accruing accuracy than going deeper or wider, especially when width and depth starts to provide diminishing returns \cite{2}.
\vskip 0.06in
\break

\subsection{Activation Function}
\vskip 0.03in
In deep neural networks, activation functions determine the output of a neuron at particular input(s) by restricting the amplitude of the output. It aids in neural network's understanding and learning process of complex patterns of the input data. Furthermore, non-linear activation functions such as ReLU6 (Rectified Linear Units capped at 6) enable neural networks to perform complex computations using fewer neurons \cite{5}.

CondenseNeXt applies ReLU6 activation function in addition to Batch Normalization technique prior to each convolutional layer. In ReLU6, units are capped at 6 to promote an earlier learning of sparse features and to prevent a sudden blowup of positive gradients to infinity. ReLU6 activation function is defined mathematically as follows:
\vskip 0.05in
\begin{equation}
\label{eqn_5}
f(x) = min(max(0,x),6)
\end{equation}
\vskip 0.05in

\begin{table*}[t]
\centering
\caption{Comparison of Performance}
\arrayrulecolor{black}
\begin{tabular}{|l|l|c|c|c|c|} 
\hline
\rowcolor[rgb]{0.831,0.831,0.831} \multicolumn{1}{|c|}{Dataset} & \multicolumn{1}{c|}{CNN Architecture} & FLOPs (in millions) & Parameters (in millions) & \multicolumn{1}{l|}{Top-1 \% Error} & Top-5 \% Error \\ 
\hline
\rowcolor[rgb]{0.961,0.961,0.961} {\cellcolor[rgb]{0.961,0.961,0.961}} & CondenseNet & 65.81 & 0.52 & 5.31 & 0.24 \\ 
\hhline{|>{\arrayrulecolor[rgb]{0.961,0.961,0.961}}->{\arrayrulecolor{black}}-----|}
\rowcolor[rgb]{0.961,0.961,0.961} \multirow{-2}{*}{{\cellcolor[rgb]{0.961,0.961,0.961}}CIFAR-10} & CondenseNeXt~ ~~ & \textbf{26.35} & \textbf{0.18} & \textbf{4.79} & \textbf{0.15} \\ 
\hline
\multirow{2}{*}{CIFAR-100~ ~ ~~} & CondenseNet & 65.85 & 0.55 & 23.35 & 6.56 \\ 
\cline{2-6}
 & CondenseNeXt~ ~ ~~ & \textbf{26.38} & \textbf{0.22} & \textbf{21.98} & \textbf{6.29} \\ 
\hline
\rowcolor[rgb]{0.961,0.961,0.961} {\cellcolor[rgb]{0.961,0.961,0.961}} & CondenseNet & 529.36 & 4.81 & 26.2 & 8.30 \\ 
\hhline{|>{\arrayrulecolor[rgb]{0.961,0.961,0.961}}->{\arrayrulecolor{black}}-----|}
\rowcolor[rgb]{0.961,0.961,0.961} \multirow{-2}{*}{{\cellcolor[rgb]{0.961,0.961,0.961}}ImageNet} & CondenseNeXt & \textbf{273.16} & \textbf{3.07} & \textbf{25.8} & \textbf{7.91} \\
\hline
\end{tabular}
\vskip 0.06in
Table I provides a comparison between CondenseNet (the baseline architecture) vs. CondenseNeXt (our ultra-efficient deep neural network architecture) in terms of performance each utilizing the training setup and infrastructure as outlined in section 6 and 7 in this paper.
\end{table*}

\section{Cyberinfrastructure}
\medskip
\subsection{Training Infrastructure}
\vskip 0.03in
\begin{itemize}
\setlength\itemsep{0.35em}
\item Intel Xeon Gold 6126 12-core CPU with 128 GB RAM.
\item NVIDIA Tesla V100 GPU.
\item CUDA Toolkit 10.1.243.
\item PyTorch version 1.1.0.
\item Python version 3.7.9.
\end{itemize}

\medskip
This cyberinfrastructure for training is provided and managed by the Research Technologies division at the Indiana University which supported our work in part by Shared University Research grants from IBM Inc. to Indiana University and Lilly Endowment Inc. through its support for the Indiana University Pervasive Technology Institute \cite{8}.

\subsection{Testing Infrastructure}
\vskip 0.03in
\begin{itemize}
\setlength\itemsep{0.35em}
\item NXP BlueBox 2.0 ARM-based autonomous embedded  development  platform.
\item Intempora RTMaps Remote Studio version 4.8.0.
\item CIFAR-10, CIFAR-100 and ImageNet Datasets.
\item PyTorch version 1.1.0.
\item Python version 3.7.9.
\end{itemize}
\medskip

\begin{figure}
\centering
\begin{tikzpicture}
\begin{axis}[
    legend pos=north west,
    axis lines=middle,
    extra x ticks=0,
    xmax=11,
    xmin=-11,
    ymin=0,
    ymax=7.5,
    xlabel={$Input$},
    ylabel={$Output$}]
\addplot [domain=-11:11, samples=500, thick, blue] {max(0,x)};
\addlegendentry{ReLU}
\addplot [domain=-11:11, samples=1000, thick, red] {min(max(0,x),6)};
\addlegendentry{ReLU6}
\end{axis}
\end{tikzpicture}
\caption{Difference between ReLU and ReLU6 activation functions.}
\end{figure}
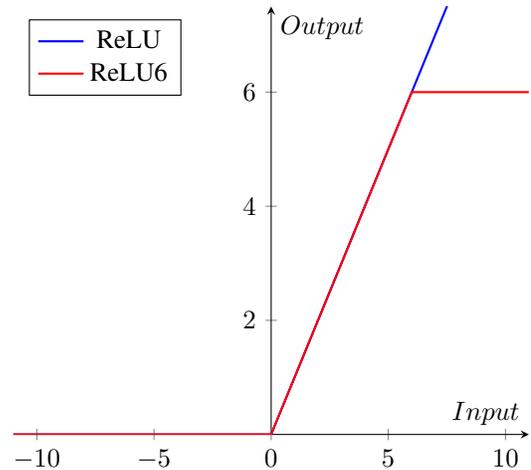

\section{Experiment and Results}
\medskip
Training results presented in this report are based on the evaluation of image classification performance of CondenseNeXt CNN on three benchmarking datasets: CIFAR-10, CIFAR-100 and ImageNet. CondenseNeXt was designed and developed in PyTorch framework and trained on NVIDIA's Tesla V100 GPU with standard data augmentation scheme \cite{1}, Nesterov Momentum Weight of 0.9, Stochastic Gradient Descent (SGD), cosine shape learning rate and dropout rate of 0.1 for all three datasets discussed in this section.

\subsection{CIFAR-10 Classification}
\vskip 0.03in
CIFAR-10 dataset \cite{4,5} was first introduced by Alex Krizhevsky in \cite{17}. It is one of the most widely used datasets for evaluating a CNN in the field of deep learning research. There are 60,000 RGB images of 10 different classes of size 32$ \times $32 pixels divided into two sets of 50,000 for training and 10,000 for testing.

CondenseNeXt was trained with a single crop of inputs on CIFAR-10 dataset for 200 epochs, batch size of 64 and features \textit{$k$} of 8-16-32. Using RTMaps Remote Studio, an image classification script was developed using Python scripting language and evaluated on NXP BlueBox for single image classification analysis. Table I provides a comparison of performance between CondenseNet and CondenseNeXt CNN in terms of FLOPs, parameters, and Top-1 and Top-5 error rates. Figure 5 provides a screenshot of the RTMaps console.

\subsection{CIFAR-100 Classification}
\vskip 0.02in
CIFAR-100 dataset was also first introduced by Alex Krizhevsky in \cite{17} along side CIFAR-10 dataset. It is also one of the many popular choices of datasets in the field of deep learning research. Just like CIFAR-10 dataset, there are 60,000 RGB images in total. However, it has 100 different classes, where each class contains 600 images of size 32$ \times $32 pixels divided into two sets of 50,000 for training and 10,000 for testing. CIFAR-100 classes are mutually exclusive of CIFAR-10 classes. For example, CIFAR-100's baby, chimpanzee and rocket classes are not part of the CIFAR-10 classes.

CondenseNeXt was trained with a single crop of inputs on CIFAR-100 dataset for 600 epochs, batch size of 64 and features \textit{$k$} of 8-16-32. Using RTMaps Remote Studio, an image classification script was developed using Python scripting language and evaluated on NXP BlueBox for single image classification analysis. Table I provides a comparison of performance between CondenseNet and CondenseNeXt CNN in terms of FLOPs, parameters, and Top-1 and Top-5 error rates. Figure 6 provides a screenshot of the RTMaps console.

\subsection{ImageNet Classification}
\vskip 0.02in
ImageNet was introduced by an AI researcher Dr. Fei-Fei Li along with a team of researchers at a 2009 IEEE Conference on Computer Vision and Pattern Recognition (CVPR) in Florida \cite{7}. This dataset is built according to the WordNet hierarchy where each node in the hierarchy corresponds to over five hundred images. In total, there are over 14 million images in this dataset that have been hand-annotated and labelled by the team.

CondenseNeXt was trained with a single crop of inputs on the entire ImageNet dataset for 120 epochs with a Group Lasso rate of 0.00001, batch size of 256, features of 8-16-32-64-128 and four Nvidia V100 GPUs using Data Parallelism technique. An image classification script was developed in RTMaps Remote Studio and evaluated on NXP BlueBox for single image classification analysis. Table I provides a comparison of performance between CondenseNet and CondenseNeXt CNN in terms of FLOPs, parameters, and Top-1 and Top-5 error rates. Figure 7 provides a screenshot of the RTMaps console.

\begin{figure}
    \centering
    \includegraphics[scale=0.307]{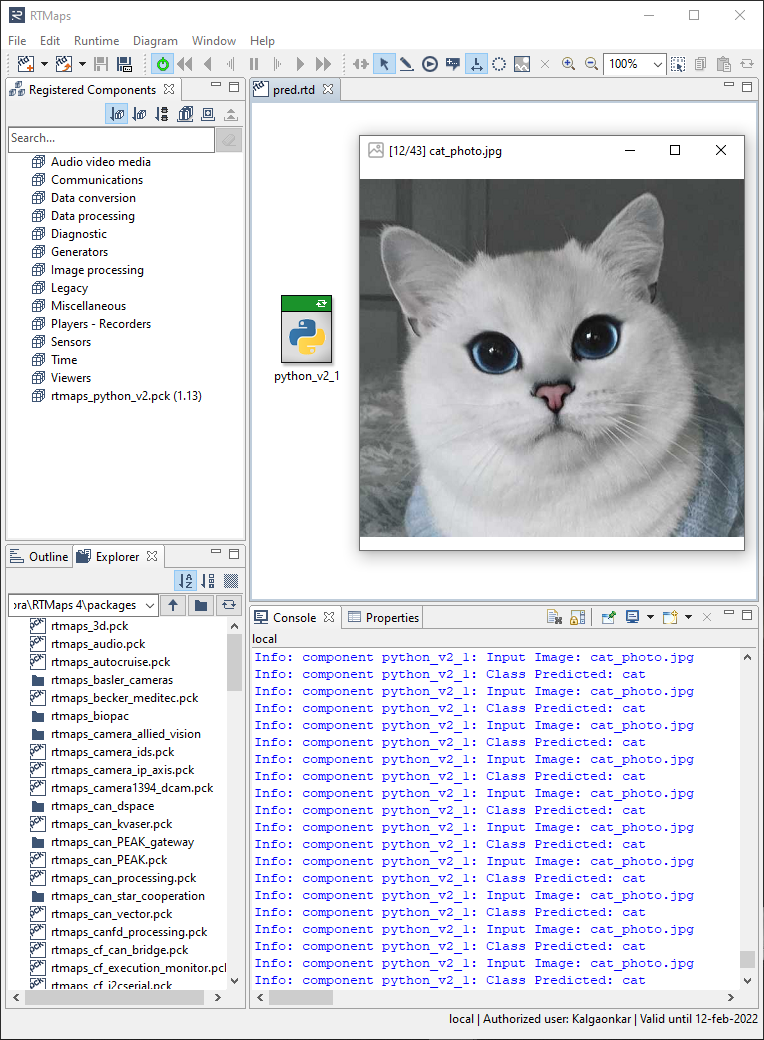}
    \caption{Evaluation of CondenseNeXt on CIFAR-10 dataset when deployed on NXP BlueBox 2.0 using RTMaps Remote Studio 4.8.0 for classifying an image of a cat and outputting the predicted class in RTMaps console.}
    \label{fig:my_label3}
\end{figure}

\begin{figure} [htbp]
    \centering
    \includegraphics[scale=0.307]{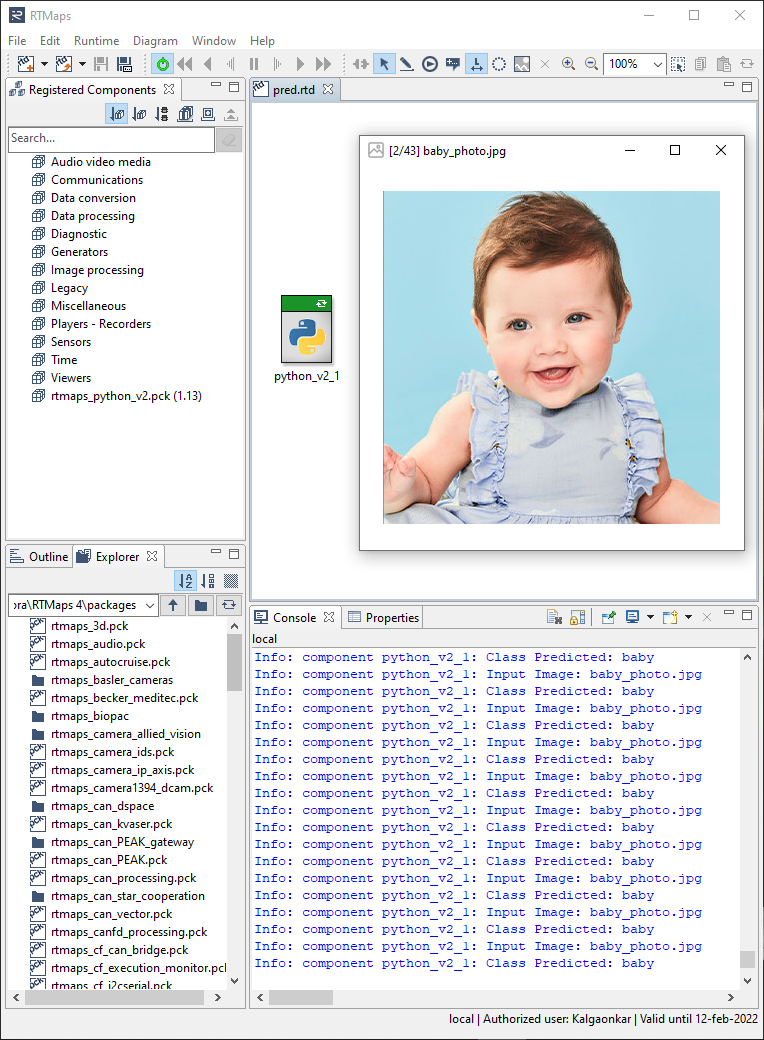}
    \caption{Evaluation of CondenseNeXt on CIFAR-100 dataset when deployed on NXP BlueBox 2.0 using RTMaps Remote Studio 4.8.0 for classifying an image of a baby and outputting the predicted class in RTMaps console.}
    \label{fig:my_label4}
\end{figure}

\begin{figure}
    \centering
    \includegraphics[scale=0.3998]{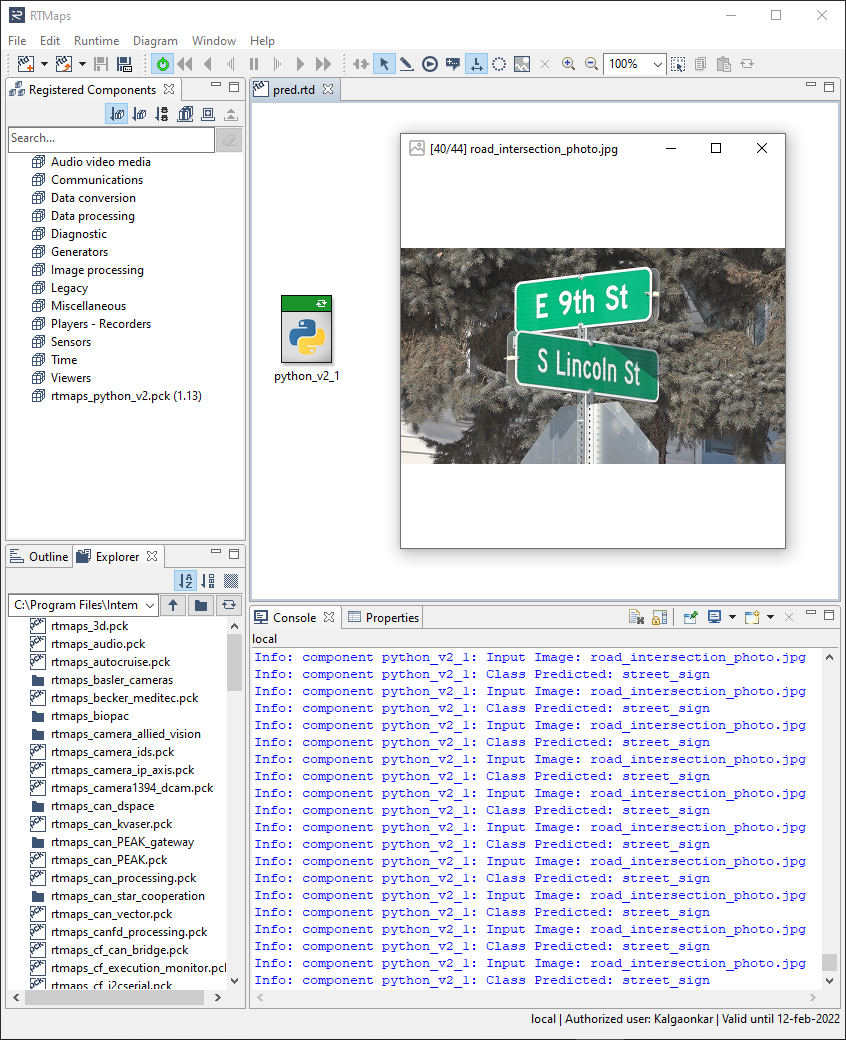}
    \caption{Evaluation of CondenseNeXt on ImageNet dataset when deployed on NXP BlueBox 2.0 using RTMaps Remote Studio version 4.8.0 for classifying an image of a street sign and outputting the predicted class in the RTMaps console.}
    \label{fig:my_label5}
\end{figure}

\section{Conclusion}
In this paper, we demonstrate the performance of CondenseNeXt CNN which is an ultra-efficient deep convolutional neural network architecture for ARM-based embedded computing platforms without CUDA enabled GPU(s). Extensive training from scratch and analysis have been conducted on three benchmarking datasets: CIFAR-10, CIFAR-100 and ImageNet. It achieves state-of-the-art image classification performance on CIFAR-10 dataset with a 4.79\% Top-1 error rate, on CIFAR-100 dataset with a 21.98\% Top-1 error rate and ImageNet dataset with a 7.91\% single model and single crop Top-5 error rate. Our experiments on NXP's BlueBox further validate the effective use Depthwise Separable Convolutional layers and Model Compression techniques implemented to discard inconsequential elements and to reduce FLOPs without affecting overall performance of the neural network. In the future, we will explore different applications with CondenseNeXt such as image segmentation and object detection to better exploit different opportunities for OpenCV applications.

\section*{Acknowledgment}
The authors would like to acknowledge the support of Indiana University Pervasive Technology Institute for providing supercomputing and storage resources that have contributed to all research results reported within this paper.

\end{document}